\documentclass[letterpaper]{article} 
\usepackage{aaai23}  
\usepackage{times}  
\usepackage{helvet}  
\usepackage{courier}  
\usepackage[hyphens]{url}  
\usepackage{graphicx} 
\usepackage{natbib}  
\usepackage{caption} 
\usepackage{boldline, makecell}
\usepackage{multirow}
\usepackage{subcaption}
\usepackage{algorithm}
\usepackage{algorithmic}

\usepackage{newfloat}
\usepackage{listings}
\DeclareCaptionStyle{ruled}{labelfont=normalfont,labelsep=colon,strut=off} 
\floatstyle{ruled}
\newfloat{listing}{tb}{lst}{}
\floatname{listing}{Listing}
\title{An Emotion-guided Approach to Domain Adaptive Fake News Detection using Adversarial Learning (Student Abstract)}
\author{
    Arkajyoti Chakraborty\equalcontrib \textsuperscript{\rm 1}, Inder Khatri\equalcontrib \textsuperscript{\rm 1}, Arjun Choudhry\equalcontrib \textsuperscript{\rm 1}, Pankaj Gupta\textsuperscript{\rm 1}, Dinesh Kumar Vishwakarma\textsuperscript{\rm 1}, Mukesh Prasad\textsuperscript{\rm 2}
}
\affiliations{
    \textsuperscript{\rm 1} Biometric Research Laboratory, Delhi Technological University, New Delhi, India\\
    \textsuperscript{\rm 2} School of Computer Science, FEIT, University of Technology Sydney, Sydney, Australia\\
    \{arkajyotichakraborty\_2k19ep022, dinesh\}@dtu.ac.in, 
    \{inderkhatri999, choudhry.arjun\}@gmail.com, mukesh.prasad@uts.edu.au
}

\usepackage{bibentry}

\begin{document}

\maketitle

\begin{abstract}
Recent works on fake news detection have shown the efficacy of using emotions as a feature for improved performance. However, the cross-domain impact of emotion-guided features for fake news detection still remains an open problem. In this work, we propose an emotion-guided, domain-adaptive, multi-task approach for cross-domain fake news detection, proving the efficacy of emotion-guided models in cross-domain settings for various datasets.
\end{abstract}

\section{Introduction}
Over the years, our reliance on social media as an information source has increased, leading to an exponential increase in the spread of \emph{fake news}. To counter this, researchers have proposed various approaches for fake news detection (FND). Models trained on one domain often perform poorly on datasets from other domains due to the domain shift (Figure 1(1)). Some works show the efficacy of domain adaptation for cross-domain FND by extracting domain-invariant features (Figure 1(2)) for classification \citep{DA_FND}. However, adapting domains does not ensure that features in different classes align correctly across domains, which sometimes has a negative impact on performance. Some works have shown a correlation between fake news and their intrinsic emotions \citep{Dual_1, Emo_FND_AAAI} (Figure 1(3)), having successfully used it for fake news detection. However, these works are restricted to in-domain settings and don't consider cross-domain evaluation. We propose the use of emotion-guided multi-task models for improved cross-domain fake news detection, experimentally proving its efficacy, and present an emotion-guided domain adaptive approach for improved cross-domain fake news detection by leveraging better feature alignment across domains due to the use of emotion labels (Figure 1(4)).

\begin{figure}[t!]
     \centering
     \includegraphics[width = 0.68\columnwidth]{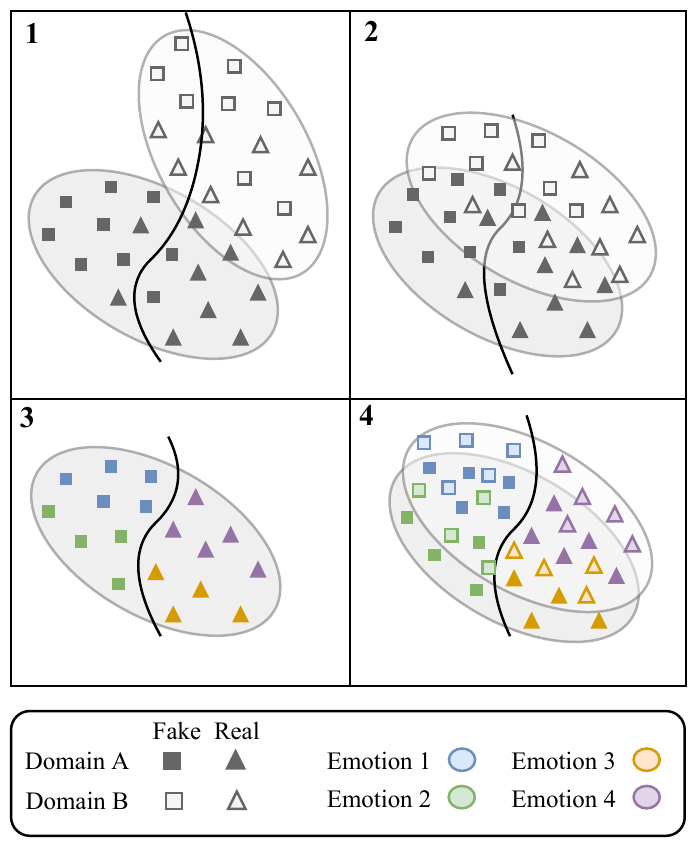}
     \hfill
   \caption{(1) Cross-domain texts not aligned. (2) Domain adaptation for improved alignment. (3) Emotion-guided classification. (4) Emotion-guided domain adaptation.}
   \label{DA_Flowchart} 
\end{figure}

\section{Proposed Methodology}

\subsection{Datasets, Emotion Annotation \& Preprocessing}
We use the FakeNewsAMT \& Celeb \citep{perez-rosas}, Politifact\footnotemark[1], and Gossipcop\footnotemark[2] datasets. We annotate them with the core emotions from Ekman's \citep{ekman} (6 emotions: \textit{Joy}, \textit{Surprise}, \textit{Anger}, \textit{Sadness}, \textit{Disgust}, \textit{Fear}) and Plutchik's \citep{plut} (8 emotions: \textit{Joy}, \textit{Surprise}, \textit{Trust}, \textit{Anger}, \textit{Anticipation}, \textit{Sadness}, \textit{Disgust}, \textit{Fear}) emotion theories. We use the Unison model \citep{unison} for annotating the datasets with emotion tags. During preprocessing, we convert text to lower case, remove punctuation, and decontract verb forms (eg. \textit{“I’d”} to \textit{“I would”}).

\footnotetext[1]{https://www.politifact.com/}
\footnotetext[2]{https://www.gossipcop.com/}

\begin{figure*}[t!]
     \centering
     \includegraphics[width = 0.8\textwidth]{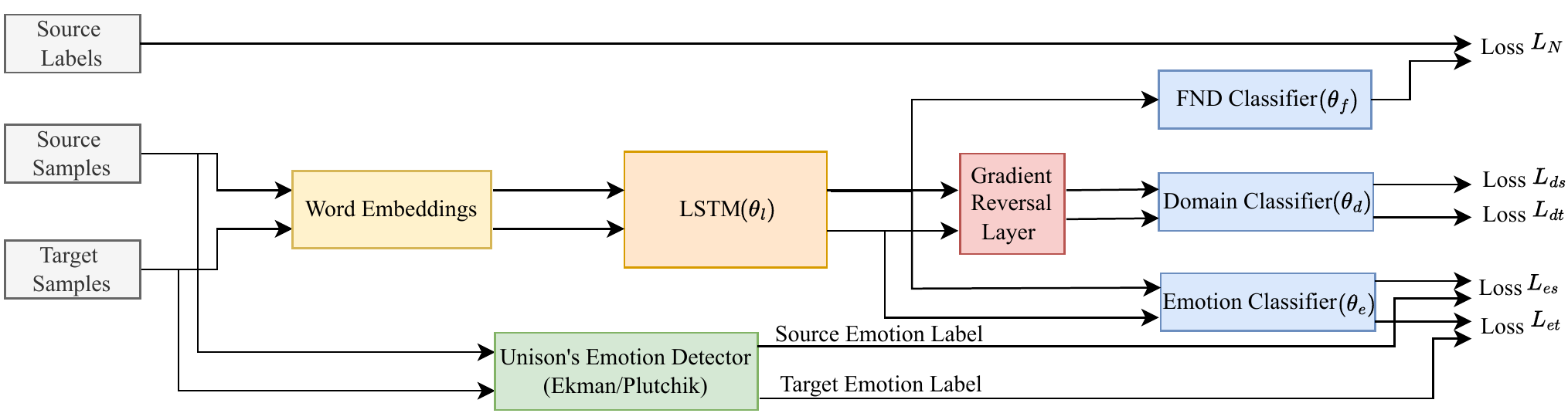}
     \hfill
   \caption{Graphical representation of our emotion-guided domain-adaptive framework for cross-domain fake news detection.}
   \label{Flowchart} 
\end{figure*}

\subsection{Emotion-guided Domain-adaptive Framework}
We propose the cumulative use of domain adaptation and emotion-guided feature extraction for cross-domain fake news detection. Our approach aims to improve the feature alignment between different domains using adversarial domain adaptation by leveraging the correlation between the emotion and the veracity of a text (as shown in Figure \ref{DA_Flowchart}(4)). Figure \ref{Flowchart} shows our proposed framework. We use an LSTM-based multi-task learning (MTL) feature extractor which is trained by the cumulative losses from fake news classifier, emotion classifier, and discriminator (aids in learning domain-invariant features). LSTM can be replaced with better feature extractors. We use it specifically for easier comparison to non-adapted emotion-guided and non-adapted single-task models. The domain classifier acts as the discriminator. Fake news classification loss, emotion classification loss, adversarial loss, and total loss are defined as:

\begin{equation}
 \small
 \scriptstyle L_{FND} \ = \ \min\limits_{\theta_{l},\theta_{f}} \sum_{i=1}^{n_{s}} L_{f}^i
\end{equation}

\begin{equation}
 \small
 \scriptstyle L_{emo} \ = \ \min\limits_{\theta_{l},\theta_{e}} \sum_{i=1}^{n_{s}} L_{es}^i \ + \ \sum_{j=1}^{n_{t}} L_{et}^j))
\end{equation}

\begin{equation}
 \small
 \scriptstyle L_{adv} \ = \ \min\limits_{\theta_d} (\max\limits_{\theta_l}( \sum_{i=1}^{n_{s}} L_{ds}^i \ + \ \sum_{j=1}^{n_{t}} L_{dt}^j))
\end{equation}
\begin{equation}
 \small
 \scriptstyle L_{Total} \ = \ (1 - \alpha - \beta) * L_{FND} \ + \ \alpha \ * \ (L_{adv}) \ + \ \beta \ * \ (L_{emo}) 
\end{equation}

where $n_s$ and $n_t$ are number of samples in source and target sets; $\theta_d$, $\theta_f$, $\theta_e$, and $\theta_l$ are parameters for discriminator, fake news classifier, emotion classifier, and LSTM feature extractor; $L_{d_s}$ and $L_{d_t}$ are binary crossentropy loss for source and target classification; $L_{es}$ and $L_{et}$ are crossentropy loss for emotion classification; $L_f$ is binary crossentropy loss for Fake News Classifier; $\alpha$ and $\beta$ are weight parameters in $L_{Total}$. We optimized $\alpha$ and $\beta$ for each setting.

\section{Experimental Results \& Discussion}
Each model used for evaluation was optimized on an in-domain validation set. Table \ref{FAMT-Celeb} illustrates our results proving the efficacy of using emotion-guided models in non-adapted cross-domain settings. Table \ref{DA} compares non-adaptive models, domain adaptive models, and our emotion-guided domain adaptive models in various settings. MTL (E) and MTL (P) refer to emotion-guided multi-task frameworks using Ekman's and Plutchik's emotions respectively. STL refers to single-task framework. DA refers to domain-adaptive framework with a discriminator. Non-DA refers to a non-adapted model. Some findings observed are:

\textbf{Emotions-guided non-adaptive multi-task models outperform their single-task counterparts in cross-domain settings}, as seen in Table \ref{FAMT-Celeb}, indicating improved extraction of features that are applicable across different datasets.

\textbf{Emotion-guided domain-adaptive models improve performance in cross-domain settings.} Table \ref{DA} shows the advantage of emotion-guided adversarial domain-adaptive models over their non-adaptive counterparts. This shows the scope for improved feature extraction even after adversarial adaptation, and emotion-guided models act as a solution. 


\begin{table}[t!]
    \centering
    \large
    \resizebox{\columnwidth}{!}{
    \begin{tabular}{c|c|c|c|c}
    \hline
    \hline
        \textbf{Source} & \textbf{Target} & \textbf{\makecell{Accuracy\\Non-DA STL}} & \textbf{\makecell{Accuracy\\Non-DA MTL(E)}} & \textbf{\makecell{Accuracy\\Non-DA MTL(P)}} \\\hline
        FAMT & Celeb & 0.420 & 0.520 & \textbf{0.530}\\
        \hline
        Celeb & FAMT & 0.432 & 0.471 & \textbf{0.476}\\
    \hline
    \hline
    \end{tabular}}
    \caption{Cross-domain evaluation of non-adaptive models on FakeNewsAMT (FAMT) \& Celeb datasets. Emotion-guided models (MTL (E) and MTL (P)) outperform their corresponding STL models in cross-domain settings.}
    \label{FAMT-Celeb}
\end{table}

\begin{table}[t!]
    \centering
    \large
    \resizebox{\columnwidth}{!}{
    \begin{tabular}{c|c|c|c|c|c}
    \hline
    \hline
        \textbf{Source} & \textbf{Target} & \textbf{\makecell{Accuracy \\ Non-DA \\ STL }} & \textbf{\makecell{Accuracy \\ DA \\ STL }}& \textbf{\makecell{Accuracy \\ DA \\ MTL(E)}} & \textbf{\makecell{Accuracy \\ DA \\ MTL(P)}}\\\hline
        FAMT & Celeb & 0.420 & 0.560 & 0.540 & \textbf{0.600}\\
        \hline
        Celeb & FAMT & 0.432 & 0.395 & 0.501 & \textbf{0.551}\\
        \hline
        Politi & Gossip & 0.527 & 0.585 & \textbf{0.698} & 0.671\\
        \hline
        Celeb & Gossip & 0.488 & 0.525 & 0.555 & \textbf{0.587}\\
        \hline
        FAMT & Gossip & 0.451 & 0.790 & \textbf{0.805} & 0.795\\
        \hline
        FAMT & Politi & 0.363 & 0.621 & \textbf{0.704} & 0.621\\
    \hline
    \hline
    \end{tabular}}
    \caption{Cross-domain evaluation of non-adaptive, adaptive and emotion-guided adaptive models on various datasets.}
    \label{DA}
\end{table}

\fontsize{9pt}{10pt}\selectfont {\bibliography{main.bib}}

\end{document}